# Fast and Scalable Multi-Robot Deployment Planning Under Connectivity Constraints*

Yaroslav Marchukov and Luis Montano[1] [†‡]

## Abstract

In this paper we develop a method to coordinate the deployment of a multi-robot team to reach some locations of interest, so-called primary goals, and to transmit the information from these positions to a static Base Station (BS), under connectivity constraints. The relay positions have to be established for some robots to maintain the connectivity at the moment in which the other robots visit the primary goals. Once every robot reaches its assigned goal, they are again available to cover new goals, dynamically re-distributing the robots to the new tasks. The contribution of this work is a two stage method to deploy the team. Firstly, clusters of relay and primary positions are computed, obtaining a tree formed by chains of positions that have to be visited. Secondly, the order for optimally assigning and visiting the goals in the clusters is computed. We analyze different heuristics for sequential and parallel deployment in the clusters, obtaining sub-optimal solutions in short time for different number of robots and for a large amount of goals.

## 1 Introduction

The usage of robots in environment inspection or monitoring becomes more frequent in critical missions. A possible application is the CBRNE defense (Chemical, Biological, Radiological and Nuclear), where robotic agents can be used to prevent hazardous situations, such as localization of dangerous objects. So the agents must be deployed to reach some locations of interest as fast as possible to detect possible threats. This is the purpose of the present work. We develop a method to coordinate a team of robots to monitor environments with obstacles and under connectivity constraints, reaching some locations of interest, which we name as primary goals. The mission of the robots is to transmit the information to a static Base Station (BS) from these positions, see Fig.1(a). This means that the robots have to be able to communicate with a BS at least at the goal locations, the rest of the time the robots being free to remain disconnected. In the present work we denote this kind of communication as intermittent. We consider limited line-of-sight (LoS) communication, commonly used in the field of robotics [1]. As in many cases the direct communication between the BS and the agents, when they are at the goal locations, is impossible, because of the limited communication range of the wireless sensors and the obstacles.

Therefore, the first problem to tackle is how to establish a mobile robotic network to transmit the information from the goal locations to the BS. The method represents a network as a tree where its nodes are the goal position, where some robots will act as relays and others will be associated to the primary goals. Let explain this with the simple example of Figs.1-2. The limited communication range of the robots as well as the obstacles in the scenario in Fig.1(a), will determine the number of robots needed as relays in each branch. The relay goals of each chain in the branches, altogether with the primary goals that connect the chain form a cluster of goals, Fig.1(b), to which a subset of robots, or all of them, will be sent.

Here the problem of how to visit the different clusters of goals arises. Some robots must reach the relay goals, just after that, their teammates can visit the primary goals of the cluster to send the information to the BS, Fig.2. We propose and analyze different ways to visit the clusters, considering the distance to move the robots between clusters, the number of required robots, and the time to visit the goals of each cluster.

The proposed approach requires two steps:

*This is the author's version of a paper accepted at the 2019 IEEE International Conference on Autonomous Robot Systems and Competitions (ICARSC). The final version is published by IEEE and is available at: 10.1109/ICARSC.2019.8733637.

†This research has been funded by project DPI2016-76676-R-AEI/FEDER-UE and by research grant BES-2013-067405 of MINECO-FEDER, and by project Grupo DGA-T45-17R/FSE

‡1 Instituto de Investigación en Ingeniería de Aragón (I3A),University of Zaragoza, Spain `yamar@unizar.es`, `montano@unizar.es`

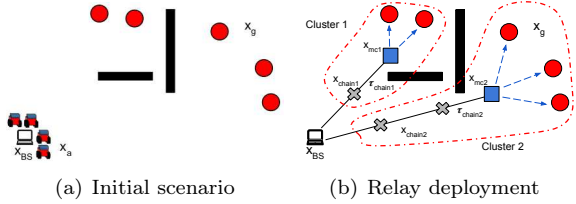

(a) Initial scenario   (b) Relay deployment

Figure 1: Relay deployment illustration. In (a) the initial scenario is illustrated: the positions of the BS, the robots and the goals to reach $\mathbf{x}_g$, depicted with red circles. (b) depicts the process of relay positions computation, where two chains of relay robots must be deployed to reach all the goals. The relays to maximize the connectivity $\mathbf{x}_{mc}$ are blue squares, the links of communication are blue dashed arrows. The chain paths $\tau_{chain}$ are depicted with black lines and the relays $\mathbf{x}_{chain}$ are gray crosses.

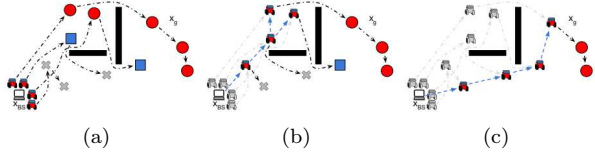

(a)   (b)   (c)

Figure 2: Order of cluster visit. (a) initial allocation, (b) visiting the first cluster, (c) visiting the second cluster. The black dashed arrows depict the allocated goals to the robots and the travelled paths. The blue dashed arrows represent the communication links between the agents.

- *Tree computing and clustering* (Fig.1): to deploy relay chains in the tree branches, and group them with the primary goals forming clusters, minimizing the number of robots devoted as relays. Explained in Sect.4.

- *Ordering the visits to the clusters* (Fig.2): to obtain the order to reach the positions in the clusters and the goals belonging to them, by minimizing the time spent in the cluster. Detailed in Sect.5.

## 2 Related Works

The deployment of a team of robots is directly related to the communication type between the agents. Two types of communication can be distinguished: permanent or intermittent. The permanent one, [2] [3] [4] [5], prevents the disconnections between agents in the course of the mission, by restricting the agents motions. It is frequently used in fast intervention missions in disaster scenarios [6]. The intermittent one, is more frequent for non-critical missions, allowing disconnections between agents and being more flexible in the agents movements. In gathering [7] [8] or patrolling [9] [10] missions, the agents exchange data with low frequency, only when they have data to share.

Exploration works [11] [12], are similar in spirit to our problem. The team also uses intermittent communication between the robots. The agents form a multi-hop net of relay robots to interconnect the BS with the explorer robots, to transmit the information. But the motions when the robots are in movement are not restricted by communication.

In [11], the algorithm finds out at every exploration step the best relay positions, using Integer Linear Programming (ILP) and a *LoS* signal model. The method guarantees the communication between the BS and explorer robots. However, the search of feasible relay positions with ILP is computationally high, because the number of possible candidates is large.

In [1], the authors analyze four different discretizations and propose an heuristic approach to reduce the search space, using the Mutual Visibility Graph (MVG). However, their heuristics evaluate different relay topologies, which is still time consuming. We propose a faster method to compute the relay locations, significantly reducing the number of potential candidates for this purpose.

After the relay positions computation, the team needs to know the order to visit these goals altogether with the primary ones (clusters). In [1], the agents visit 9 waypoints with communication with the BS, which can be solved even by brute force. In the exploration in [11], the goals lay over the frontier between the explored and unexplored areas. Thus, the number of goals is not very high, although it increases with the new explored areas. In our proposed approach we test different heuristics to deploy the robots to visit hundreds of goals reducing the mission time.

In our previous work [13] we dealt with a similar problem, but the solutions were obtained for a low number of robots and goals. As new contributions here, the developed method provides solutions for a much larger number of robots and goals, also overcoming the limitations described in the above mentioned works.

This work develops: 1) a fast method to obtain the relay positions to reach the primary goals with connection with BS, via multi-hop network; 2) a formulation and evaluation of different heuristics to visit large amounts of goals clusters in sequential or concurrent manner. The combination of both points,

**Algorithm 1** Deployment general procedure
___
**Require:** $grid, x_{BS}, \mathbf{x}_a, \mathbf{x}_g$
1: $\mathbf{x}_{mc} \leftarrow max\_con\_relays(grid, x_{BS}, \mathbf{x}_g)$ ▷ Sect.4.1
2: $\mathbf{x}_{chain} \leftarrow relay\_chains(grid, x_{BS}, \mathbf{x}_{mc})$ ▷ Sect.4.2
3: $\mathbf{x}_{cl} \leftarrow \mathbf{x}_{mc} \cup \mathbf{x}_{chain} \cup \mathbf{x}_{pg}$ ▷ Form clusters (Sect.4)
4: $\mathbf{x}_{cl}^* \leftarrow visit\_order(grid, \mathbf{x}_a, \mathbf{x}_{cl})$ ▷ Sect.5.2
5: $<\mathbf{x}_a, \mathbf{x}_{cl}^*> \leftarrow allocation(grid, \mathbf{x}_a, \mathbf{x}_{cl}^*)$ ▷ Sect.5.1
6: $\tau \leftarrow compute\_paths(grid, \mathbf{x}_a, \mathbf{x}_{cl}^*)$
___

provides a low time-consuming method scalable to the number of robots and goals.

## 3 System overview

### 3.1 Problem setup

The scenario is modeled as a grid, because the *Fast Marching Method*, described below, is used in this work. The robots move in a scenario with static obstacles, where $x$ and $\mathbf{x}$ denote a position and a set of positions in the grid, respectively. There are $M$ robots located every instant at $\mathbf{x}_a = [x_{a_1}, ..., x_{a_M}]$. The mission of the robots is to reach $N$ goal locations $\mathbf{x}_g = [x_{g_1}, ..., x_{g_N}]$, and to transmit the information from the goals to the BS located at the $x_{BS}$. In this work we consider $M \leq N$. Each agent is equipped with a wireless sensor to communicate with the rest of the team, and a *line-of-sight* (LoS) communication model within a distance range $d_\gamma$ is used. The agents must adopt a chain formation from the primary goals positions to the BS. Then they have to obtain the visit order of the chains, reducing the total time of the mission.

### 3.2 Proposed approach

The general procedure of the proposed method is described in Alg.1. Firstly, the positions providing maximum connectivity where to connect the maximum number of primary positions are obtained, denoted as $\mathbf{x}_{mc}$ in l.1. With this property the minimal number of robots for relay tasks are devoted, in such a way that most of them are used to visit the primary goals.

From these positions the relay robots will provide connectivity to their teammates to visit the primary goals and to transmit the information to the BS. Due to $d_\gamma$, most of $\mathbf{x}_{mc}$ points are outside the communication area of the BS. Therefore, chains of relay goals $\mathbf{x}_{chain}$ from $x_{BS}$ to each $\mathbf{x}_{mc}$ have to be formed, in l.2. We group the relay chain goals $\mathbf{x}_{chain}$, the maximum connectivity positions $\mathbf{x}_{mc}$ and the primary goals $\mathbf{x}_{pg}$ connected from $\mathbf{x}_{mc}$, forming different clusters of goals $\mathbf{x}_{cl}$ in l.3. After the clustering process, there exist different ways to visit the goals with communication. First, the algorithm allocates the best visit order of clusters to fulfill the mission in minimal time, in l.4. Then, the cluster goals are allocated to be visited by the agents, taking into account the type of the goals, primary or relays, in l.5. Finally, the algorithm computes the shortest paths to visit all the goals, in l.6.

### 3.3 Fast Marching Method (FMM)

The deployment method is based on the use of the FMM [14] for different tasks: path planning, relay chain deployment, and evaluation of the costs for the allocation algorithms. FMM propagates a wavefront from a start position over every point of the grid, computing the distance gradient $\nabla D$ to every point, as can be seen in Fig.3(b). The wavefront propagates uniformly in all the directions with a velocity $F$, avoiding the obstacles. $F$ is 0 in cells that contain an obstacle and is 1 in a free space cell. Descending this gradient $\nabla D$, the direct path is obtained to the origin of the wavefront, the red line in Fig.3(e). If we propagate the wavefront from the obstacles, we obtain the distances to the closest obstacle, $\nabla D_{obst}$ Fig.3(c). The FMM is applied again from the same source, but fixing $F = \nabla D_{obst}$ the resulting gradient $\nabla D'$ of Fig.3(d) considers the distance to the obstacles. The path obtained by descending this gradient is maintained away from the obstacles, depicted with blue line in Fig.3(e). As can be seen in the image, the resulting path corresponds to the positions of the Voronoi Graph (VG) of these area. This path computed from BS, hereinafter referred to as Voronoi Path (VP), is used in our method to obtain the possible candidate positions where to deploy the relay chains, as will be described in Sect.4.2.

Furthermore, we consider advantageous FMM for goals allocation to robots. With a unique gradient computation we know the distances to all the positions of the grid. So, the allocation algorithms used in this work employ FMM to analyze the costs to assign the goals to the robots, as will be described in Sect.5.

## 4 Relay chains and clustering

In the present section we describe the computation of the relay locations to visit the primary goals with communication with the BS.

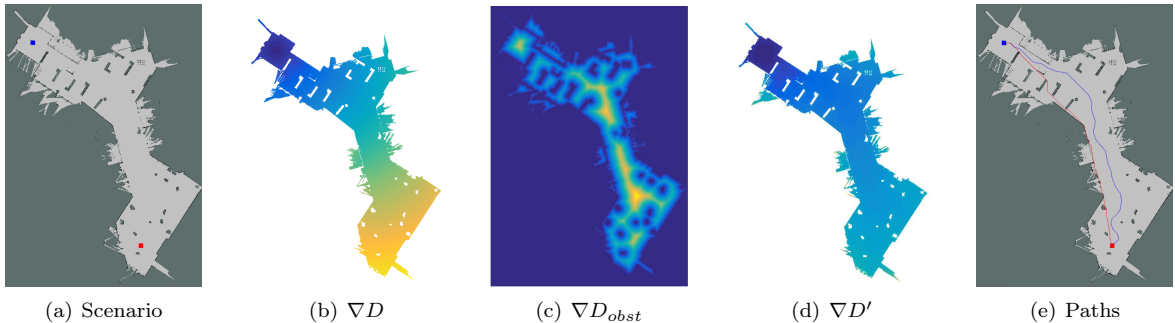

| (a) Scenario | (b) $\nabla D$ | (c) $\nabla D_{obst}$ | (d) $\nabla D'$ | (e) Paths |

Figure 3: Path computation with FMM. In (a), blue and red squares are start and goal positions respectively. (b) gradient from start. (c) gradient from obstacles. (d) gradient with $F = \nabla D_{obst}$. (e) red path obtained from $\nabla D$ of (b), blue path (VP) descending $\nabla D'$ in (d).

### 4.1 Maximum Connectivity Relays

As mentioned in Sect.3.2, the idea is to devote the minimal number of robots for relay tasks during the mission. Thus, the maximum possible number of robots is used to visit the primary goals. For this purpose, we use the Minimum Steiner Tree (MST), since it is the best option for connectivity maximization, as it has been demonstrated in [1] [11] [12].

First, we need to find out the possible candidate positions where placing the relays to reach the primary goals with connectivity. The algorithm obtains all the possible positions where it will be possible to establish a communication link with the primary goals, by computing all the positions that satisfy the *LoS* communication model.

Then, the algorithm iteratively computes the positions of maximum connectivity, $\mathbf{x}_{mc}$. Firstly, the algorithm discards the primary goals within the communication range of the BS, because no relays are needed to reach these goals. The visit of the remaining primary goals will require the use of relays. At each iteration, the algorithm selects the position where there are more communication links from more not yet connected primary goals. If there are multiple positions that provide the same connectivity, the method chooses the one closest to the BS, which a priori reduces the number of required relay robots and their displacements. The algorithm iterates until connect all the primary goals. As depicted in Fig.1(b), the algorithm firstly obtains $x_{mc_2}$, because it has communication links with 3 primary goals. Then it computes $x_{mc_1}$, to provide connectivity to the remaining two goals. The obtained $\mathbf{x}_{mc}$ correspond to the points of MST because they are the minimum necessary to interconnect the primary goals with the BS.

### 4.2 Branches of Relay Chains

As stated in Sect.2, the main drawback of the relay placement is the high computational cost to analyze the suitable candidates for this purpose. In [1], the authors evaluate four different discretizations of the environment to obtain the possible candidates for relay positions: two polygonal discretizations, a grid of a desired resolution, and points of the Voronoi Graph (VG). The VG provides a good balance between the cost of the solutions and the execution time. This is because the potential candidates to place the relays are those positions that are distant from the obstacles, with wider field of view, and the number of points to evaluate is low, in comparison to evaluate the entire grid as in [11] [12].

At the same time, it is logical that the relay chain will be placed somewhere along or close to the direct path from the BS to the goal for *LoS* communication. Therefore we use the paths from $x_{BS}$ to each $\mathbf{x}_{mc}$ to place the chain relays. We use the Voronoi Paths (VP), defined in Sect.3.3. The advantage of using VP with respect to compute the entire VG, is that the path will guide the relay chain directly to the desired $x_{mc}$, instead of computing the complete VG. At the same time, VP only considers the useful points of VG to the desired $x_{mc}$, instead of considering the points of the entire graph. Since VP contains fewer points than VG, the search of the relay positions is faster.

After obtaining all the VPs connecting $\mathbf{x}_{mc}$ with $x_{BS}$, the algorithm computes the positions of the relays on these paths, placing the relay positions with an interval $d_\gamma$ and with *LoS* along the path from $x_{mc}$ to $x_{BS}$.

An illustrative example of the procedure of connection of primary goals and clusterization is depicted in Fig.1(b). For two $\mathbf{x}_{mc}$, two chains $\tau_{chain}$ are extended. The resultant configuration has two clusters composed by: primary goals, maximum connectivity relay $x_{mc}$ and the relay chain goals $x_{chain}$. This approach minimizes the movements of the chains; once the robots acting as relays are deployed, they remain in their position until their teammates finish the visit of primary goals in the cluster.

## 5 Goal and cluster allocation

The proposed approach employs two allocation procedures: the allocation of relays and primary goals to the agents within the clusters, and the order in which the clusters have to be visited.

### 5.1 Relay and Primary Goals Allocation

The robots sequentially visit the relay and primary goals in different clusters, using the Hungarian method to optimally allocate the goals. It assigns a set of works to a set of workers, in such a way that the sum of the working costs is the minimal. In our problem the distance between robots and goals positions are the working costs. The distances from every goal to the agents are computed by the FMM, being the value of the gradient at the position of the agent the cost to reach the goal for this agent. A distance matrix $D$, which is used by the Hungarian method to allocate the goals, is computed.

Generally, the number of goals in the clusters is higher than the number of robots. The relay goals have priority over the primary ones, because without deploying the relays, the robots that visit the primary goals cannot transmit data. In order to ensure this order, we modify the costs in matrix $D$ corresponding to relay goals with the expression:

$$D^*_{relay} = \frac{D_{relay} \cdot min(D)}{max(D_{relay})} \quad (1)$$

where $D_{relay}$ are the values of the matrix that correspond only to the distances to the relay goals, $min(D)$ is the global minimum of the matrix $D$, and $max(D_{relay})$ denotes the maximum distance to the relay goals. This way, the costs that correspond to reach the relay goals, $D^*_{relay}$, are always lower than the costs of primary goals; thus the Hungarian algorithm always allocates first the relay goals to the agents.

After this allocation with priorities, the team is able to extend a relay chain and it must have at least one agent to start visiting the primary goals of the cluster. We also use the Hungarian method to iteratively assign the primary goals to the robots that does not belong to the relay chains. In the present work we assume a precise continuous localization for the agents. The collision avoidance during the navigation towards the goals is out of the scope of this paper. An attitude control for aerial robots or a reactive collision avoidance for ground robots would be applied during the execution.

### 5.2 Cluster Visit Order

In this section we develop different heuristics to visit the different goals clusters, in sequential and concurrent manner. The use of heuristic approaches is motivated by the need to be able to allocate the tasks to the agents in a short time, few seconds, against classic orienteering problem methods, which take minutes to find a solution [15].

#### 5.2.1 Sequential Cluster Visit

For this type of cluster visit we evaluate different costs based on the distances between clusters, the amount of required robots and the workload in each cluster. The costs are the following:

- *Distance between clusters (CD)*. This cost measures the displacements between the relay chains. The distance matrix is obtained using the distances between maximum connectivity relay positions of the clusters:

$$CD = \|x_{mc_i} - x_{mc_j}\|, i = 1, ..., K, j = 1, ..., K, i \neq j \quad (2)$$

where $K$ denotes the number of clusters. Again, FMM is used to obtain the distances. The solution is provided solving the Travelling Salesman Problem (TSP). As we want to obtain the solution as fast as possible, we employ different solvers based on the number of instances of the problem, empirically adjusted for our computer. If the number of instances is at most 12, we use the brute force, obtaining the optimal solution. For instances greater than 12 and lower than 20, we use the branch and bound method, and for greater instances we employ the Nearest Neighbor initialization altogether with the local optimization using 2-opt method [16]. It consists in a local optimization of the route, swapping every two edges of the route, goals in our case, checking if the cost of the new route improves the previous one. The algorithm selects the different routines in order to obtain a solution within an interval time of 50 milliseconds.

- *Required relays (RA)*, obtained as:

$$RA = (M_{rcl_i} + 1)/N, \quad i = 1, ..., K \quad (3)$$

where $M_{rcl}$ denotes the number of relays of the cluster and one agent to visit the primary goals, $N$ is the number of robots. With this procedure the robots visit the clusters based on the required relays in ascendant order. This approach can be interesting in corridor-like environments with rooms.

- *Mean cost of the cluster*:

$$CMD = \overline{\|x_{mc_i} - \mathbf{x}_{p_i}\|}, \quad i = 1, ..., K \quad (4)$$

where $\mathbf{x}_{p_i}$ are the positions of the primary goals of cluster $i$. It measures the approximated cost of visiting the primary goals in each cluster.

- *Worst cost of the cluster*:

$$CWD = max(\|x_{mc_i} - \mathbf{x}_{p_i}\|), \quad i = 1, ..., K \quad (5)$$

The same as the previous one but it considers the worst cost for each cluster.

We evaluate different costs to obtain the sequential visit order of the clusters, listed in the Table 1. The overbar stands for the normalized values of the variables, to the maximum. The symbol $\times$ denotes the product of the proposed costs. $PA$ is the number of primary goals in the cluster, that measures the amount of workload within each cluster. Heuristic $S3$ is the combination of $S1$ and $S2$, that sorts the cluster visit based on distance from BS, and $S4 - S8$ consider the displacements and the workload within each cluster.

#### 5.2.2 Concurrent Cluster Visit

In this kind of cluster visit we propose different techniques to simultaneously extend multiple chains of relays to reach the primary goals of the clusters. We define three different ways to obtain the adjacency graph between the clusters. The BS is the root, from where the team starts, and the vertices are $\mathbf{x}_{mc}$ of the clusters. The algorithm, starts from the root, and iteratively connects the disconnected vertices to the connected ones by levels. The levels represent the number of hops from BS.

- *Relay Number Level (RL)*: connecting the vertices based on the number of required relays to reach the clusters, in ascending order. For example, for clusters that require $\{3, 5, 1, 0, 1, 5\}$ relays, the algorithm will sort and connect the clusters obtaining as result $\{0, 1, 1, 3, 5, 5\}$.

- *Cluster Distance Level (DL)*: connecting the clusters based on the lowest distance between clusters, eq.(2). The algorithm, at each iteration, connects one disconnected vertex to the closest connected vertex.

Table 1: List of tested methods

| | Sequential | | Concurrent |
|---|---|---|---|
| S1 | $CD$ | C1 | $RL$-$LC$ |
| S2 | $RA$ | C2 | $RL$-$MC$ |
| S3 | $\overline{CD} \times \overline{RA}$ | C3 | $RL$-$MP$ |
| S4 | $\overline{CD} \times \overline{PA}$ | C4 | $DL$-$LC$ |
| S5 | $\overline{CD} \times \overline{CMD}$ | C5 | $DL$-$MC$ |
| S6 | $\overline{CD} \times \overline{CWD}$ | C6 | $DL$-$MP$ |
| S7 | $\overline{CD} \times \overline{PA} \times \overline{CMD}$ | C7 | $RDL$-$LC$ |
| S8 | $\overline{CD} \times \overline{PA} \times \overline{CWD}$ | C8 | $RDL$-$MC$ |
| | | C9 | $RDL$-$MP$ |

- *Relay and Distance Level (RDL)*: it is a mix between $RL$ and $DL$. At each iteration, the algorithm selects one non-connected vertex from the next relay level, obtained with $RL$, and connects it to the closest connected vertex.

After obtaining the graph of clusters, the team distributes the robots between them, extending several chains. We propose three different strategies of chain extensions:

- *LC*: extending all the possible chains only to the next level.

- *MC*: extending the maximum number of possible chains.

- *MP*: extending the chains needed to visit the maximum number of primary goals.

With $LC$ and $MC$ criteria, the agents that will visit the primary goals are distributed proportionally to the number of primary goals of the reachable clusters. Reachable clusters are those that have already extended relay chains. Thus, the evaluated concurrent strategies are the combinations of the aforementioned techniques, listed in the Table 1.

## 6 Evaluation

We test our method in simulation for different situations and for different $\#robots/\#goals$ ratios. The results are evaluated with respect to the total mission time. The scenario is shown in Fig.3(a), with dimensions $45.8 \times 65.65$m and a resolution of 20 cm. The goals are generated from an uniform distribution. The velocity for all the robots is $0.2m/s$. The communication range is $d_\gamma = 10m$.

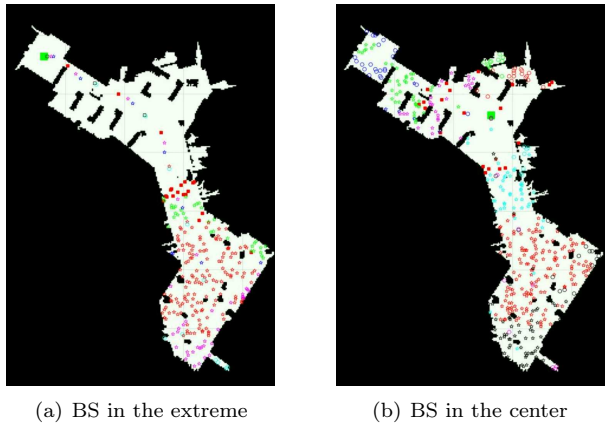

(a) BS in the extreme  (b) BS in the center

Figure 4: Snapshot of the tested scenarios. Big green rectangle is the BS, red squares are the agents and the rest of the markers are clusters of goals.

## 6.1 Scalability and influence of the BS position

We test the strategies for $N = [5, 10, 20, 50]$ robots and $M = [100, 200, 500]$ primary goals and for two different positions of BS, Fig.4: one in an extreme area of the scenario and another one centered. In Fig.4(b) there are more possibilities to extend multiple relay chains in different directions, to execute in parallel the visits to clusters. As the purpose of this work is to reach the goals with connectivity for large teams of robots and number of goals, the avoidance between agents is not considered. We run 10 simulations for randomly distributed goals for each scenario. The mean results of total mission time for the heuristics of Table 1 are depicted in Fig.5.

With 5 robots, only 50% and 90% of the goals are reached in Fig.4(a) and Fig.4(b), respectively. Thus, there cannot be a fair comparison with the cases with larger robots instances, in which all the goals are reached.

The influence of the BS position is observed in Fig.5(a),5(b),5(c). For Fig.4(a), it can be seen a better performance of the sequential methods, because there are very few possibilities to extend several chains, parallelizing the cluster visit. Only for large teams (50) the concurrent methods outperform the sequential ones. When the BS is centered, Fig.4(b), the teams of few robots (10) fulfill the mission in similar time using sequential and some of the concurrent methods, specially with $LC$ heuristics. For larger teams (20 and 50), there is a significant improvement of the concurrent approaches, since there are enough agents to extend several chains in different directions.

From the sequential methods, in general, the approaches $S1,S3,S6$ provide better results. All of them consider the distance between clusters including $CD$ cost. $S3$ also forces the robots to visit first the clusters with smaller number of relays, giving priority to the clusters closer to the BS. $S6$, using $CWD$ cost of eq.(5), penalizes the clusters with more dispersed goals or where the goals are within areas difficult to access. A common case in our scenario due to its layout and $LoS$ communication model.

Regarding to the concurrent methods, the best methods to sort the clusters visits are $RL$ and $RDL$, because they take into account the relays to reach the clusters, which implicitly include the distances from the BS. The better results are provided specially using $LC$ heuristics, in $C1,C4,C7$. It is because these techniques deploy chains in different directions, but prioritize the clusters closer to the BS. We see a clear improvement of the best concurrent techniques with respect to the best sequential ones for 20 and 50 robots, being 30% and 21%, respectively in Fig.5(e). In Fig.5(f) it is 17% and 31%, for 20 and 50 robots respectively. The concurrent methods C3,C6,C9 ($MP$ heuristics) extend the chains to visit the maximum number of primary goals. This causes a sequential and oscillatory behaviour, very counterproductive when there are many robots. The team extends only one chain and the robots go from one side of the map to another.

We can conclude that the most reliable of the sequential methods for the different BS positions and for different number of goals is S1 ($CD$), which only considers the distance between the clusters to order the visits. When the size of the robot team increases, the concurrence provides better results, particularly when the BS is centered, extending several chains to visit different clusters of goals. Being C1 ($RL\text{-}LC$) the method that, in mean, obtains the better results for all #robots/#goals ratios.

## 6.2 Computation Time

The simulations were implemented on C++ and performed on a machine with Intel Core i7-4770 processor clocked at 3.4GHz with 8GB of RAM. Our algorithm proceeds in two steps: first, computation of the positions of the relay goals, and then, the allocation of the visit order of the goals. The $\mathbf{x}_{mc}$ are computed to maximize the primary goals covered from each relay position, from the intersection of their communication areas. The time to compute the communication area of one primary goal is $8ms$. Then, the VP is computed from the BS to each $\mathbf{x}_{mc}$, requiring two FMM gradient computations, one from the ob-stacles $\nabla D_{obst}$ of Fig.3(c), and another from the BS, $\nabla D'$ of Fig.3(d). Both gradients require $50ms$ in total. Then, the path is obtained descending $\nabla D'$ in less than $1ms$. Note that the gradients are computed once for all the mission and the paths VP are computed for every $\mathbf{x}_{mc}$. Placing the relay goals over one VP takes $7ms$.

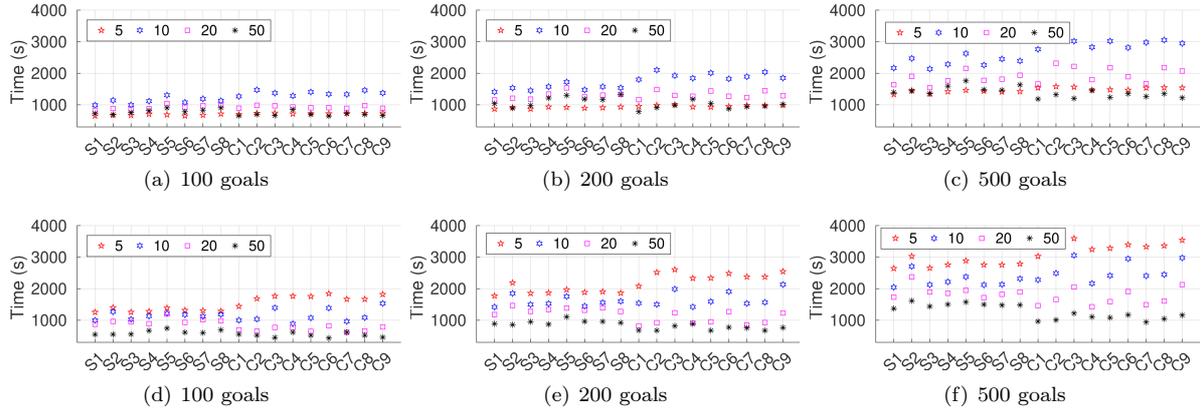

Figure 5: Total time of the mission. Fig.(a)-(c) are for the BS in a extreme position. Fig.(d)-(f) are for the BS in the center of the scenario.

Table 2: Mean computation times of relay and cluster computation and allocation, expressed in seconds

|  | Relay goals | Allocation |  |  |  |
| --- | --- | --- | --- | --- | --- |
|  |  | 5 | 10 | 20 | 50 |
| **100** | 1.72 | 0.23 | 0.62 | 0.8 | 1.52 |
| **200** | 2.67 | 0.4 | 1.01 | 1.28 | 2.23 |
| **500** | 5.67 | 0.96 | 2.14 | 2.6 | 4.1 |

The computation times to obtain the relay positions and to allocate the goals to the robots are shown in the Table.2. The proposed relay computation is faster than some works in the literature with similar deployment objectives. In [1], a solution to deploy 8 robots to visit 9 waypoints is obtained in minutes. In [11], 12 robots explore the environment. The objectives for the robots lay over the frontiers between explored and unexplored areas, being at most a couple of dozens of goals. The solution is found in several seconds. Our approach is able to obtain a solution for these instances of $\#robots/\#goals$, in less than a second in the worst case. The worst case would imply that each primary goal is within an independent cluster.

## 7 Conclusions

We have presented a method to deploy a team of robots to visit a high number of locations of interest and to transmit the information to a static base station, under connectivity constraints. It is a fast and scalable method to compute the relay positions to reach the goals with connectivity, which improves the computation time with respect to other techniques for similar objectives in the literature. The approach groups the goals into different clusters in order to obtain sub-optimal solutions of visit order for high number of goals. We have proposed and evaluated several sequential and concurrent heuristics to visit the clusters of the goals, in order to obtain those that result in the shortest times of the mission. We can conclude that the sequential approaches are more efficient for low ratios $\#robots/\#goals$ and for the BS located in extreme positions in the scenario. However, for high ratios the concurrent routines reduce the mission time, deploying different relay chains of robots. For future works we will generalize our method for both intermittent and permanent connectivity. Since our method is focused on large teams of robots and many goals, we also want to include the bandwidth constraint as in [12], to avoid the latency problems in the real-world communication between the robots.